\documentclass[letterpaper, 10 pt, conference]{ieeeconf}  
\IEEEoverridecommandlockouts                              

\usepackage{amssymb}
\usepackage{amsmath}
\usepackage{bbm}
\usepackage{subcaption}
\usepackage{graphics} 

\PassOptionsToPackage{hyphens}{url}\usepackage[breaklinks=true,hidelinks=true,colorlinks=true,citecolor=blue,linkcolor=blue]{hyperref}

\usepackage[ruled,vlined,linesnumbered]{algorithm2e}
\usepackage{algpseudocode}
\usepackage{bm}
\usepackage{enumerate}
\usepackage{graphicx}
\usepackage{lipsum}
\usepackage{multirow}
\usepackage{tabularx}
\usepackage{framed}
\usepackage[dvipsnames]{xcolor}

\usepackage{pgfplots}
\usepackage{tikz}
\usepackage{booktabs}
\usetikzlibrary{positioning}
\usetikzlibrary{calc, arrows, fit, spy}

\definecolor{codegreen}{rgb}{0.0,0.6,0.0}
\usepackage{tikz}
\pgfplotsset{compat=1.18}
\usetikzlibrary{pgfplots.groupplots}
\usetikzlibrary{shapes.geometric, arrows, positioning, fit, calc}

\makeatletter

\newcolumntype{Y}{>{\centering\arraybackslash}X}

\definecolor{bblue}{HTML}{4F81BD}
\definecolor{rred}{HTML}{85c1e9}
\definecolor{ggreen}{HTML}{27ae60}
\definecolor{ppurple}{HTML}{884ea0}

\newcommand{\ignore}[1]{}

\title{\LARGE \bf
CropTrack: A Tracking with Re-Identification\\ Framework for Precision Agriculture
}

\author{Md Ahmed Al Muzaddid$^{*}$, Jordan A. James$^{*}$, and William J. Beksi
\thanks{$^{*}$ Indicates equal contribution. This work was supported by a
University of Texas at Arlington Dissertation Fellowship and by the United
States Department of Agriculture (USDA) under agreement \#58-6066-3-050.}
\thanks{The authors are with the Department of Computer Science and
        Engineering, The University of Texas at Arlington, Arlington, TX, USA.
        Emails:
        mdahmedal.muzaddid@mavs.uta.edu,
        jaj9608@mavs.uta.edu,
        william.beksi@uta.edu.
        }
}

\begin{document}

\maketitle
\pagestyle{plain}

\begin{abstract}
Multiple-object tracking (MOT) in agricultural environments presents major
challenges due to repetitive patterns, similar object appearances, sudden
illumination changes, and frequent occlusions. Contemporary trackers in this
domain rely on the motion of objects rather than appearance for association.
Nevertheless, they struggle to maintain object identities when targets undergo
frequent and strong occlusions. The high similarity of object appearances makes
integrating appearance-based association nontrivial for agricultural scenarios.
To solve this problem we propose CropTrack, a novel MOT framework based on the
combination of appearance and motion information. CropTrack integrates a
reranking-enhanced appearance association, a one-to-many association with
appearance-based conflict resolution strategy, and an exponential moving average
prototype feature bank to improve appearance-based association. Evaluated on
publicly available agricultural MOT datasets, CropTrack demonstrates consistent
identity preservation, outperforming traditional motion-based tracking methods.
Compared to the state of the art, CropTrack achieves significant gains in
association accuracy and identification precision scores with a lower number of
identity switches. 
\end{abstract}


\section{Introduction}
\label{sec:introduction}
\begin{figure}[t]
\centering
\resizebox{\columnwidth}{!}{
    \pgfplotsset{
        compat=1.3,
    }

\begin{tikzpicture}
    \begin{axis}[
        width=4.0in,
        height=3.5in,
        grid=major,
        major grid style={line width=.25pt,draw=gray!25},
        xlabel=AssA,
        ylabel=IDP,
        xtick distance=5,
        scatter/classes={%
                DeepSORT={draw=blue, fill=blue},
                ByteTrack={draw=teal, fill=teal},
                PineSORT={draw=yellow, fill=yellow},
                FastTracker={draw=orange, fill=orange},
                NTrack={draw=blue, fill=blue},
                AgriSORT={draw=green, fill=green},
                CropTrack={draw=red, fill=red}}
    ]
        \addplot[
            scatter=true,
            only marks,
            mark=*,
            scatter src=explicit symbolic,
            visualization depends on={.003*\thisrow{HOTA}*\thisrow{HOTA} \as \perpointmarksize},
            scatter/@pre marker code/.append style={
                /tikz/mark size=\perpointmarksize,
            },
            nodes near coords*={\Label},
            visualization depends on={value \thisrow{label} \as \Label},
            visualization depends on={value \thisrow{Pos} \as \myPos},
            visualization depends on={\thisrow{addOffset} \as \myOffset},
            visualization depends on={\thisrow{xshift} \as \myXShift}, 
            visualization depends on={\thisrow{IDP} \as \myval},
            every node near coord/.append style={
                font=\small,
                \myPos=\perpointmarksize pt + \myOffset pt,
                xshift=\myXShift pt 
            },
        ] table [y={IDP},x={AssA},meta=label] {
            HOTA    IDP    AssA    label           anchor  Pos    addOffset xshift
            44.59   76.41   50.78   {ByteTrack}     south   left   0         0
            45.81   72.44   46.44   {PineSORT}      south   left  0         0
            46.58   68.98   50.13   {FastTracker}   south   below  0        -5
            24.74   40.39   17.08   {NTrack}        south   above  0         0
            43.90   50.68   46.09   {AgriSORT}      south   left  0         0
            46.04   79.05   53.35   {CropTrack}     south   above  -3        -3
        };
  \end{axis}
\end{tikzpicture}}
\caption{A comparison of state-of-the-art trackers on the AgriSORT-Grapes
\cite{saraceni2024agrisort} dataset. The horizontal axis is the association
accuracy (AssA) score, the vertical axis is the identification precision (IDP),
and the radius of each circle corresponds to the higher-order tracking accuracy
(HOTA) score. CropTrack achieves the best AssA and IDP score while maintaining
comparable HOTA performance.}
\label{fig:bubble_plot}
\vspace{-2mm}
\end{figure}
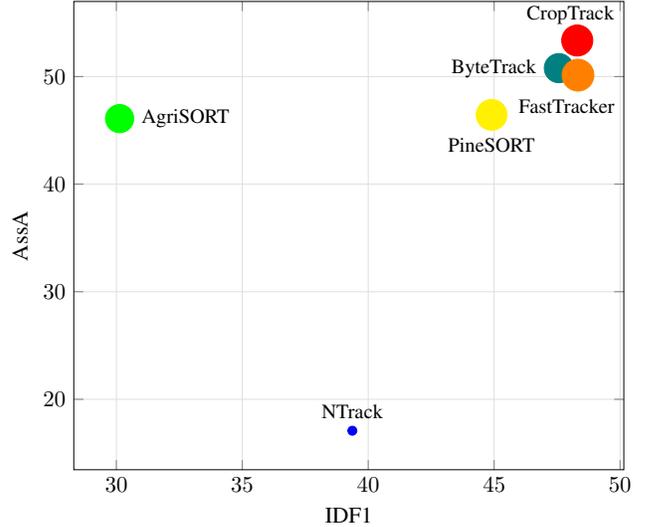

Driven by labor shortages, robotics is emerging as a pivotal technology to
address the need for more efficient and sustainable food production. For
example, robotic platforms are being developed for tasks such as precision
spraying, mechanical weeding, crop monitoring, and automated harvesting
\cite{botta2022review}. A fundamental prerequisite for these systems is the
ability to sense and perceive complicated and dynamic environments causally,
ensuring that detections are handled as they occur. At the core of this
perceptual capability lies multiple-object tracking (MOT), a computer vision
task focused on detecting and maintaining the identities of many objects across
a sequence of video frames.

Accurate MOT is necessary to achieve advanced perception required for
agricultural robotics. For instance, tracking plants and weeds over time enables
the precise application of fertilizers or herbicides, minimizing chemical usage
and environmental impact. Similarly, accurate detection and tracking of
individual fruits can allow a robotic harvester to plan an optimal picking
trajectory and avoid redundant harvesting attempts. MOT is also essential for
yield estimation \cite{villacres2023apple}, where tracking individual crops can
provide precise counts. Thus, the ability to maintain a persistent track of each
object of interest is integral for the efficiency and reliability of
agricultural automation.

Applying MOT in agricultural settings presents a formidable set of challenges.
Frequent occlusions caused by foliage, branches, or other objects are common,
leading to fragmented tracks and identity switches. Furthermore, the objects of
interest typically exhibit high intra-class similarity and are often found in
dense clusters, making it difficult to distinguish between individual instances.
The appearance of crops can also change dramatically due to variations in
lighting and shadows. To overcome these issues, it is crucial to develop MOT
frameworks specifically tailored to agricultural. Such methods must
differentiate visually similar objects, handle severe occlusions, and be capable
of operating in an online manner. 

In this letter we propose CropTrack, a re-identification (Re-ID) MOT framework
designed for complex agricultural environments. Our approach combines
Re-ID-based association using reranking of appearance features and motion-based
association to improve data association and maintain object identities through
long-term occlusions, Fig.~\ref{fig:bubble_plot}. By overcoming these
limitations, our work aims to enhance the perceptual capabilities of robots for
precision agriculture. To summarize, we make the following contributions.
\begin{itemize}
  \item We create a MOT system that incorporates appearance-based association
  specifically designed for video applications with long-term occlusions.
  \item We develop a one-to-many motion-based association strategy with greedy
  appearance-based conflict resolution.
  \item We introduce an efficient reranking technique to refine appearance-based
  association.
\end{itemize}
The source code and multimedia associated with this project can be found at
\href{https://robotic-vision-lab.github.io/croptrack}{https://robotic-vision-lab.github.io/croptrack}.

\section{Related Work}
\label{sec:related_work}
\subsection{Tracking by Detection}
\label{subsec:tracking_by_detection}

Kalman filters are widely used for location prediction due to their simplicity
(e.g., \cite{li2010multiple,cao2023observation}). However, the assumption of
linear dynamics and Gaussian noise limits their effectiveness for complex
motion. Researchers have explored more flexible approaches such as particle
filters, extended Kalman filters, and nonparametric interpolation methods (e.g.,
Gaussian process regression (GPR)). On the other hand, deep learning techniques
such as POI \cite{yu2016poi} and Tracktor \cite{bergmann2019tracking}, use a
convolutional neural network (CNN) to generate discriminative embeddings. These
embeddings are essential for preserving identities during long-term occlusions
\cite{wang2022recent}. 

Simple online and real-time tracking (SORT) \cite{bewley2016simple} combines
Kalman filtering with the Hungarian algorithm \cite{kuhn1955hungarian} and is
one of the earliest tracking-by-detection approaches. Subsequent extensions,
including DeepSORT \cite{wojke2017simple} and StrongSORT
\cite{du2023strongsort}, incrementally improve tracking performance. In
particular, DeepSORT augments SORT by incorporating appearance features, while a
Kalman filter with the Mahalanobis distance supports short-term motion-based
predictions. Building on DeepSORT, StrongSORT introduces a spatio-temporal
connectivity model for tracklet association, GPR-based interpolation to address
trajectory gaps caused by missing detections, and exponential moving average
(EMA) updates of appearance features. ByteTrack \cite{zhang2022bytetrack}
extends the association logic by employing low-confidence detection boxes to
maintain trajectory continuity during partial occlusions. This approach is
further enhanced by FastTracker \cite{hashempoor2025fasttracker}, which
introduces an occlusion-aware Re-ID module that restores lost tracks using
motion-based geometric reasoning rather than traditional appearance-based
features.

Recent research increasingly focuses on end-to-end or joint
detection-association architectures. In these networks, detection and tracking
cues are learned simultaneously rather than in separate modules. For example,
TransTrack \cite{sun2020transtrack}, TrackFormer
\cite{meinhardt2022trackformer}, and MOTR \cite{zeng2022motr} employ
transformer-based query and key mechanisms to jointly detect objects in the
current frame and associate them with existing tracks within a unified pipeline.
Such methods reduce dependence on post-hoc association heuristics, but they rely
on large amounts of labeled spatial-temporal datasets to learn robust
association functions.

\subsection{Appearance-Based Re-Identification}
\label{subsec:appearance-based_re-identification}
Appearance-based Re-ID has been investigated in domains such as pedestrian and
vehicle tracking, where the objective is to consistently match object instances
across non-overlapping camera views or over time despite variations in
appearance, pose, and illumination. Early pedestrian Re-ID approaches rely on
handcrafted descriptors, such as color histograms and texture features, to
encode visual information \cite{liao2015person}. CNN-based methods enable the
learning of highly-discriminative representations
\cite{zheng2016person,hermans2017defense}. Proceeding advancements introduce
attention mechanisms \cite{zhao2017spindle,li2018harmonious} and part-based
modeling \cite{sun2018beyond,kalayeh2018human} to emphasize salient regions and
improve cross-view generalization. 

Zhang et al. \cite{zhang2023pha} further enhance discriminative capability
through patch-wise high-frequency augmentation (PHA), which preserves critical
high-frequency components via self-attention. To resolve ambiguities among
visually similar individuals (e.g., identical clothing color or accessories),
several reranking strategies have been proposed (e.g.,
\cite{ye2015coupled,zhong2017re}) to refine similarity estimation post-matching.
Inspired by the success of these techniques in disambiguating near-identical
appearances, we incorporate reranking-based similarity-refinement into our MOT
framework. To the best of our knowledge, we are the first to adapt an
appearance-based reranking scheme for crop-tracking tasks.

\subsection{Crop Tracking}
\label{subsec:crop_tracking}
By maintaining consistent identities across video streams, crop tracking can
support a range of downstream applications. For instance, LettuceTrack
\cite{hu2022lettucetrack} introduces handcrafted geometric features that
leverage the relative positions of plants along a row, exploiting the regular
spatial arrangement of crops. While effective in highly-structured environments,
this approach lacks distinctiveness and robustness, and its reliance on
grid-like planting patterns limits broader applicability. In a related
development, AgriSORT \cite{saraceni2024agrisort} argues that appearance-based
association is often unreliable in agricultural domains where targets share
near-identical visual characteristics. The framework depends exclusively on
motion cues and assumes that robot movement remains largely parallel to orchard
rows to minimize perspective distortion. However, this assumption may not hold
under diverse field conditions. Furthermore, we show that appearance-based
reranking strategies, when combined with motion cues, can enhance association
performance despite the inherent visual similarity of crops. 

Similarly, PineSort \cite{xie2025pinesort} uses ORB \cite{rublee2011orb}
features to mitigate camera motion and adopts a multi-stage, confidence-driven
matching strategy. However, like AgriSORT, PineSort relies mainly on motion
information, thus its association accuracy decreases in scenarios where motion
cues are weak or ambiguous leading to reduced robustness. In contrast to these
approaches, NTrack \cite{muzaddid2024ntrack} models linear relationships among
neighboring tracks without assuming row- or grid-based planting. By integrating
both direct cues such as dense optical flow and indirect spatial relationships,
NTrack improves the consistency and reliability of detection-to-track
association, especially in fields with irregular planting patterns. WeedsSORT
\cite{jin2025weedssort} utilizes a multi-dimensional feature extraction decoder
and establishes keypoint correspondences via the SuperGlue
\cite{sarlin2020superglue} algorithm to derive motion estimation results. For
data association, WeedsSORT relies on template matching, which is susceptible to
erroneous matches under challenging field conditions such as occlusions, varying
illumination, or crop movement caused by wind.

\section{Method}
\label{sec:method}
\subsection{Review of ByteTrack}
\label{subsec:review_of_bytetrack}
CropTrack employs appearance-based association by integrating aspects of
ByteTrack \cite{zhang2022bytetrack} as the baseline tracker. ByteTrack is a
hierarchical framework for detection-to-track matching. Detection bounding boxes
are matched to existing tracks in multiple steps based on detection accuracy.
For each video frame $k$, detections $\mathcal{D}_{k}$ are partitioned into
high-confidence $\mathcal{D}_{high}$ and low-confidence $\mathcal{D}_{low}$ sets
using a score threshold $\tau$. A Kalman filter is then applied to predict the
positions of all tracks $\mathcal{T}$, including those that are lost. The first
association step matches $\mathcal{D}_{high}$ with predicted tracks using
intersection over union (IoU), and assignments are solved via the Hungarian
algorithm. Unmatched detections and tracks are retained as
$\mathcal{D}_{remain}$ and $\mathcal{T}_{remain}$, respectively. In the second
stage, the remaining tracks are further associated with $\mathcal{D}_{low}$,
allowing the tracker to recover potential true positives with initially lower
confidence. 

\subsection{CropTrack Overview}
\label{subsec:croptrack_overview}
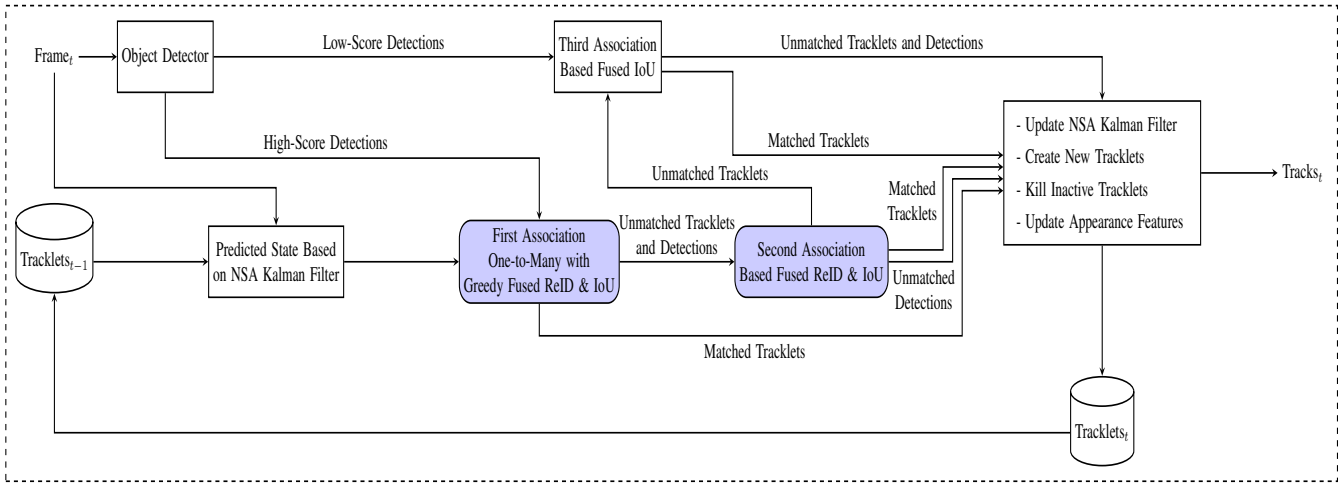
\begin{figure*}
\centering
\resizebox{\textwidth}{2.0in}{\begin{tikzpicture}[
    node distance=2.2cm and 3cm, 
    process/.style={
        rectangle, 
        draw, 
        black, 
        thick, 
        minimum height=1.2cm, 
        minimum width=2.5cm, 
        align=center
    },
    special/.style={
        process, 
        rounded corners=3mm, 
        fill=blue!20,
    },
    storage/.style={
        cylinder, 
        draw, 
        black, 
        thick, 
        shape border rotate=90, 
        aspect=0.3, 
        minimum height=1.5cm, 
        align=center,
    },
    decision/.style={
        diamond, 
        draw, 
        black, 
        thick, 
        minimum size=2.5cm, 
        align=center,
    },
    arrow/.style={
        -{stealth}, 
        thick
    },
    green_arrow/.style={
        arrow,
        green!50!black
    },
    label_style/.style={
        sloped, 
        midway, 
        font=\normalsize
    }
]

\node[storage] (tracklets_tm1) {Tracklets$_{t-1}$};
\node[above=of tracklets_tm1, xshift=0cm] (frame_in) {Frame$_t$};
\node[process, right=1cm of frame_in] (detector) {Object Detector};
\node[process, right=of tracklets_tm1]  (kalman) {Predicted State Based\\on NSA Kalman Filter};
\node[special, right=of kalman] (first_assoc) {First Association\\One-to-Many with\\Greedy Fused ReID \& IoU};
\node[process, right=of detector, xshift = 5.9cm] (third_assoc) {Third Association\\Based Fused IoU};
\node[special, right=of first_assoc] (second_assoc) {Second Association\\Based Fused ReID \& IoU};

\node[process, right=of second_assoc, yshift=1.5cm] (management) {
    \begin{tabular}{l}
        - Update NSA Kalman Filter \\
        - Create New Tracklets \\
        - Kill Inactive Tracklets \\
        - Update Appearance Features
    \end{tabular}
};

\node[storage, below=of management] (tracklets_t) {Tracklets$_t$};
\node (tracks_t) [right=of management, xshift=-1cm] {Tracks$_t$};

\draw[arrow] (frame_in) -- (detector);
\draw[arrow] (frame_in) |- (0,1.25) -| (kalman);
\draw[arrow] (detector) --  (third_assoc) node[label_style, pos=0.5, above] {Low-Score Detections};
\draw[arrow] (detector) |-  (5,1.75) node[label_style, above, pos=1.5, yshift=-.05cm] {High-Score Detections} -| (first_assoc) ;
\draw[arrow] (tracklets_tm1) -- (kalman);
\draw[arrow] (kalman) -- (first_assoc);

\draw[arrow] (first_assoc.east) -- (second_assoc.west) node[label_style, pos=0.5, above, align=center] {Unmatched Tracklets\\ and Detections};

\draw[arrow] (first_assoc.south)  |- (16,-1.25)  -| (23.75,-1.25) node[label_style, pos=0.15, below, align=center] {Matched Tracklets} |- ([yshift=-.3cm]management.west);

\draw[arrow] (second_assoc) |- (16,1.25) node[label_style, above, xshift=-2.65cm] {Unmatched Tracklets} -| (third_assoc) ;

\draw[arrow] (second_assoc.east)  -| node[label_style, pos=0.275, below, align=center] {Unmatched\\Detections} (23.5,1.4) -- ([yshift=-.1cm]management.west) ;

\draw[arrow] ([yshift=0.2cm]second_assoc.east) -| (23.25,1.6) -- ([yshift=0.1cm]management.west) node[label_style, near start, below, align=center, xshift=-1.2cm, yshift=-.1cm] {Matched\\Tracklets};

\draw[arrow] ([yshift=-.25cm]third_assoc.east) -| (17.75,2.75)  |-  ([yshift=.3cm]management.west) node[label_style, pos=0.5, above, align=center, xshift=2.25cm] {Matched Tracklets};

\draw[arrow] (third_assoc.east) -| (management.north) node[label_style, pos=0.25, above] {Unmatched Tracklets and Detections};

\draw[arrow] (management) -- (tracks_t);
\draw[arrow] (management.south) -- (tracklets_t);

\draw[arrow] (tracklets_t.west) -|  (tracklets_tm1.south);

\node[draw, dashed, thick, inner sep=0.25cm, fit=(tracklets_tm1) (detector) (tracks_t) (tracklets_t) (frame_in)] (container) {};

\end{tikzpicture}}
\caption{An overview of the CropTrack pipeline where our key contributions are
shaded in blue. The pipeline begins with an object detector that generates both
low- and high-score detections. The high-score detections are processed by the
first appearance-based association. Then, all unmatched detections and tracklets
proceed to the second appearance-based association step. Finally, unmatched
tracklets are processed with the low-score detections in the third IoU-based
association.}
\label{fig:architecture}
\end{figure*}

\begin{algorithm}
\SetAlgoLined
\DontPrintSemicolon
\SetNoFillComment
\footnotesize
\KwIn{Detected bounding boxes $\mathcal{D}$; Bounding box feature embeddings $\mathcal{F}$; detection score threshold {$\tau$}}
\KwOut{Tracks $\mathcal{T}$ of the video}
Initialization: $\mathcal{T} \leftarrow \emptyset$\;
\For{ $\mathcal{D}_k$ in $\mathcal{D}$}{
	$\mathcal{D}_{high} \leftarrow \emptyset$ \;
	$\mathcal{D}_{low} \leftarrow \emptyset$ \;
	\For{$d$ in $\mathcal{D}_k$}{
	\If{$d.score > \tau$}{
	$\mathcal{D}_{high} \leftarrow  \mathcal{D}_{high} \cup \{d\}$ \;
	}
	\Else{
	$\mathcal{D}_{low} \leftarrow  \mathcal{D}_{low} \cup \{d\}$ \;
	}
	}
    \BlankLine
	\BlankLine
	\tcc{predict new locations of tracks}
	\For{$t$ in $\mathcal{T}$}{
	$t \leftarrow \texttt{KalmanFilter}(t)$ \;
	}
    \BlankLine
    \BlankLine
	\tcc{first association}
    \textcolor{codegreen}{
    $\mathcal{T}_{match}, \mathcal{D}_{match} \leftarrow $ one\_to\_many\_reranking\_association($\mathcal{T}$, $\mathcal{D}_{high}$, $\mathcal{F}$) \;
    } \label{alg:line:first_association} 
	$\mathcal{D}_{remain} \leftarrow \mathcal{D}_{high} \setminus \mathcal{D}_{match} $ \;
	$\mathcal{T}_{remain} \leftarrow \mathcal{T} \setminus \mathcal{T}_{match} $ \;
	\BlankLine
	\BlankLine
    \tcc{second association}
	\textcolor{codegreen}{
    $ \mathcal{T}_{rematch}, \mathcal{D}_{rematch} \leftarrow$ reranking\_association($\mathcal{T}_{remain}$, $\mathcal{D}_{remain}$, $\mathcal{F}$) \;
    } \label{alg:line:second_association} 
    $\mathcal{D}_{remain} \leftarrow \mathcal{D}_{remain} \setminus \mathcal{D}_{rematch} $ \;
	$\mathcal{T}_{re-remain} \leftarrow \mathcal{T}_{remain} \setminus \mathcal{T}_{rematch}$ \;
    \BlankLine
	\BlankLine

    \tcc{third association}
    $ \mathcal{T}_{re-rematch} \leftarrow$ IoU\_association($\mathcal{T}_{re-remain}$, $\mathcal{D}_{low}$) \;
    \label{alg:line:third_association} 
    $\mathcal{T}_{re-remain} \leftarrow \mathcal{T}_{re-remain} \setminus \mathcal{T}_{re-rematch}$ \;
    \BlankLine
	\BlankLine
	\tcc{delete unmatched tracks}
	$\mathcal{T} \leftarrow \mathcal{T} \setminus \mathcal{T}_{re-remain}$ \;
    \BlankLine
	\BlankLine
	\tcc{initialize new tracks}
    \For{$d$ in $\mathcal{D}_{remain}$}{
	$\mathcal{T} \leftarrow  \mathcal{T} \cup \{d\}$ \;
	}
}
Return: $\mathcal{T}$
\caption{CropTrack}
\label{algo:croptrack}
\end{algorithm}

\begin{figure*}
\centering
\resizebox{\textwidth}{!}{\input{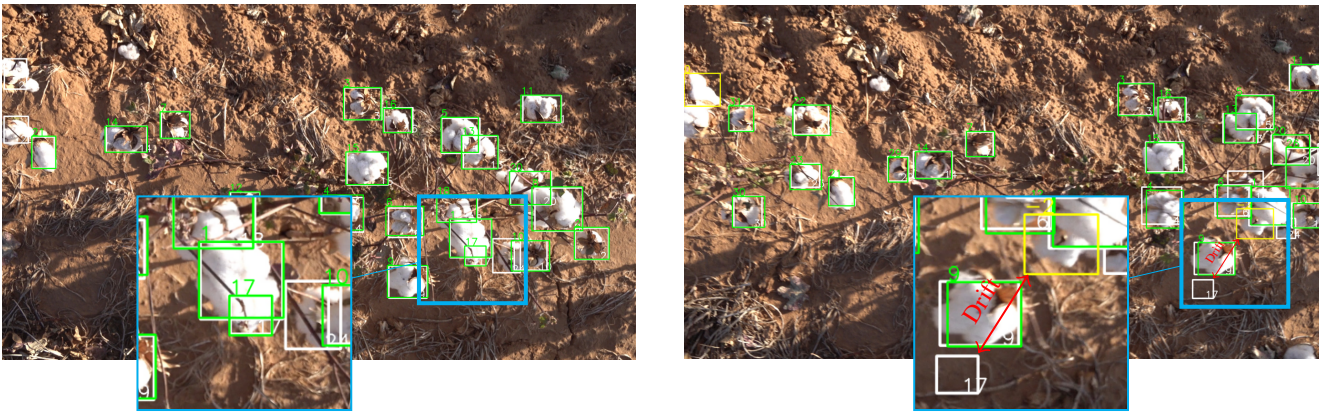}}
\caption{An example of IoU-based association failure. Left: In frame 16, a new
track is initiated with bounding box ID 17. Right: By frame 43, the track (white
bounding box, ID 17) has drifted from the object's true position, leaving the
yellow detection box unmatched and failing to associate with the correct track.}
\label{fig:iou_failure_case}
\vspace{-4mm}
\end{figure*}

CropTrack processes detected bounding boxes in a video sequence, along with
their feature embeddings and detection scores as input. As output it produces
tracks, where each track contains the bounding box and the identity of the
object across frames. Unlike ByteTrack's IoU-based association, CropTrack adopts
a more sophisticated association procedure, Fig.~\ref{fig:architecture}.
Concretely, we employ a hierarchical association strategy between detection
boxes and tracks, as listed in Algorithm~\ref{algo:croptrack}. The principal
contributions we introduce beyond ByteTrack are highlighted in
lines~\ref{alg:line:first_association} and~\ref{alg:line:second_association}.

As shown in Fig.~\ref{fig:iou_failure_case}, position prediction via a Kalman
filter can be unreliable in dynamic environments. It can lead to an accumulated
drift from the true object position, which makes an IoU-based association
ineffective. Following StrongSORT \cite{du2023strongsort}, we replace the
vanilla Kalman filter with a noise scale adaptive (NSA) Kalman filter. The NSA
Kalman filter includes the confidence of detections in the state update,
producing more robust predictions. 

In the MOT literature, appearance-based association has been proposed as a
remedy. Nonetheless, incorporating appearance cues into crop tracking is
difficult due to self-similarity, where multiple targets exhibit near-identical
visual characteristics. We solve this problem by introducing a reranking
technique to refine appearance-based association. Furthermore, we reformulate
the assignment problem from the matching cost matrix using a one-to-many
association strategy with greedy Re-ID for conflict resolution. In the ensuing
subsections we provide a detailed discussion of these key design choices. 

\subsection{Appearance-Based Association}
\label{subsec:appearance-based_association}
Enhancing MOT through appearance-based association presents a unique challenge
due to the high degree of self-similarity among targets. To fix this
complication, we adapt a reranking scheme originally developed for pedestrian
Re-ID, where it is employed to distinguish individuals with similar attributes
such as clothing color, accessories, or hairstyle \cite{zhong2017re}.
Specifically, a $k$-reciprocal encoding approach is utilized. The technique
computes a feature vector by aggregating the $k$-reciprocal nearest neighbors of
a given sample and subsequently reranks the neighbors based on a combined
distance metric that incorporates both the original distance and the Jaccard
distance.

In CropTrack, detected bounding boxes in each frame serve as the query set $Q$,
while an EMA-based feature bank
(Sec.~\ref{subsec:exponential_moving_average_feature_prototypes}) associated
with the active tracks functions as the gallery set $G$. Formally, given a query
feature vector $\mathbf{q} \in Q$, its $k$-nearest neighbors in $G$ are defined
as $N(\mathbf{q}, k) = \{\mathbf{g}_1, \mathbf{g}_2, \dots, \mathbf{g}_k\}$,
from which the $k$-reciprocal nearest neighbors are computed as $R(\mathbf{q},
k) = \{\mathbf{g}_i \mid (\mathbf{g}_i \in N(\mathbf{q}, k)) \land (\mathbf{q}
\in N(\mathbf{g}_i, k))\}$. However, variations in illumination, pose,
occlusion, and viewpoint may exclude true positives from $N(\mathbf{q}, k)$.
Zhong et al. \cite{zhong2017re} addressed this limitation by refining
$N(\mathbf{q},k)$ and computing a $k$-reciprocal nearest-neighbor distance,
$d^*(\mathbf{q}, \mathbf{g}_i)$, between all pairs of query-to-gallery samples. 

Nevertheless, detection boxes tend to associate with only spatially-local
tracks. This is based on the observation that although Kalman filter track
predictions are not accurate, the predicted tracks generally remain close to the
true object location. Let $\mathbf{c}_q \in \mathbb{R}^2$ and $\mathbf{c}_{g_i}
\in \mathbb{R}^2$ denote the 2D spatial centers of the bounding boxes
corresponding to the query embedding $\mathbf{q}$ and the gallery embedding
$\mathbf{g}_i$, respectively. This property guides us in refining the pairwise
distance,
\begin{equation}
  d^\dagger(\mathbf{q}, \mathbf{g}_i) =
  \begin{cases}
    d^*(\mathbf{q}, \mathbf{g}_i), & \text{if } \|\mathbf{c}_q - \mathbf{c}_{g_i}\|_2 < \delta \\
    \infty, & \text{otherwise}
  \end{cases}
  \label{eq:refined_pairwise_distance}
\end{equation}
where $\delta$ serves as a parameter that controls the maximum allowable spatial
distance for potential association. This formulation enforces spatial
consistency while preserving the robustness of appearance-based reranking,
ultimately improving association accuracy in highly self-similar crop tracking
scenarios.

\subsection{One-to-Many Association with Appearance-Based Conflict Resolution}
\label{subsec:one-to-many_association_with_appearance-based_conflict_resolution}
A common way to associate predicted tracklet states and current detections is to
formulate a linear assignment problem that can be solved using the Hungarian
algorithm. This assignment is based on minimizing a cost matrix derived from a
spatial metric like IoU, an appearance metric such as cosine similarity, or a
combination of the two metrics. However, this one-to-one assignment approach can
fail in highly-dynamic scenes or under partial occlusions. Global optimization
may prioritize a spatially-convenient match that does not correspond to the
correct object identity, leading to identity switches. Even when appearance
features are fused into the cost matrix, their influence is weighted against the
spatial metric across all possible assignments. This can suppress the ability to
resolve local ambiguities effectively. To address this limitation, we design a
two-stage, one-to-many motion-based association strategy with appearance-based
conflict resolution. 

The association process consists of three stages: candidate generation,
appearance feature reranking, and greedy conflict resolution. In the first
stage, a pool of potential candidates is generated. The IoU distance is computed
for all pairs of existing tracks and new detections. Any track-detection pair
with an IoU distance below a predefined threshold is considered a plausible
match and added to a candidate list. In the second stage, the reranking distance
\eqref{eq:refined_pairwise_distance} is calculated for every pair in the
candidate list. To maintain efficiency, this step is performed only on
spatially-plausible pairs, not on all possible pairs. In the final stage, the
candidate list is sorted in ascending order based on the computed distances and
then greedily processed to assign matches. The pair with the highest appearance
similarity is selected as a definitive match and removed from consideration for
further matches. This process repeats for the next-best pair until all conflicts
are resolved.

\subsection{Exponential Moving Average Feature Prototypes}
\label{subsec:exponential_moving_average_feature_prototypes}
Although the EMA-based feature bank has gained popularity in MOT due to
significant computational savings and its ability to suppress detection noise,
it is not without flaws. For example, the strategy for updating the feature bank
is greatly affected by the chosen value of momentum, $\alpha$, which represents
the sensitivity to new features. This hyperparameter requires tuning to
accurately fit a tracker to new environments. High $\alpha$ values can fail to
accurately represent features, since the detections may include features from
occluding structures. In contrast, low $\alpha$ values struggle in scenarios
where the features do not update smoothly over time due to low-frame rates or
fast camera movements. To solve this problem, we implement an EMA-based feature
bank containing a predetermined number of prototypes that updates the appearance
state $\mathbf{e}^t_{i,p}$ for the $i$-th tracklet and $p$-th prototype in frame
$t$ with varying levels of sensitivity $\alpha_p$. Concretely,
\begin{equation}
  \mathbf{e}_{i,p}^t = \alpha_p \mathbf{e}_{i,p}^{t-1} + (1 - \alpha_p)\mathbf{f}_i^t,
  \label{eq:ema_prototype_features}
\end{equation}
where $\mathbf{f}_i^t$ is the appearance feature of the current detection. For
CropTrack, the number of prototypes is set to 3 with $\alpha_p$ values of 0.1,
0.5, and 0.9 corresponding to low, medium, and high sensitivity to new feature
updates, respectively. The additional prototypes improve robustness without the
need for manual tuning.

\subsection{Motion-Based Association}
Motion information in agricultural scenarios is generally robust and leads to
strong predictions. Beyond the first association, we use the motion predictions
from the NSA Kalman filter to enhance the cost matrix for association. Formally,
we combine the appearance cost matrix $\mathbf{C}_a$ and motion cost matrix
$\mathbf{C}_m$ for all appearance-based association steps as
\begin{equation}
  \mathbf{C} = \lambda \mathbf{C}_a + (1 - \lambda)\mathbf{C}_m,
  \label{eq:cost_fusion}
\end{equation}
where $\lambda$ is a tunable weight factor set to 0.75. Lastly, the standard
second association from ByteTrack is adopted as the third association in
CropTrack. Specifically, the final association step of CropTrack uses only
motion information to associate the low-confidence detections, all unmatched
detections, and all unmatched tracklets.

\section{Evaluation}
\label{sec:evaluation}

\begin{table*}
\centering
\begin{tabularx}{\textwidth}{l|YYYYYYY|YYYYYYY}
\toprule
\multirow{2}{*}{Method} & \multicolumn{7}{c|}{TexCot22} & \multicolumn{7}{c}{AgriSORT-Grapes} \\
\cmidrule(lr){2-8} \cmidrule(lr){9-15}
                & HOTA\textcolor{teal}{$\uparrow$}  & MOTA\textcolor{teal}{$\uparrow$}  & AssA\textcolor{teal}{$\uparrow$}  & IDF1\textcolor{teal}{$\uparrow$}  & IDP\textcolor{teal}{$\uparrow$}    & IDsw\textcolor{teal}{$\downarrow$} & Frag\textcolor{teal}{$\downarrow$} & HOTA\textcolor{teal}{$\uparrow$}  & MOTA\textcolor{teal}{$\uparrow$}  & AssA\textcolor{teal}{$\uparrow$}   & IDF1\textcolor{teal}{$\uparrow$}   & IDP\textcolor{teal}{$\uparrow$}   & IDsw\textcolor{teal}{$\downarrow$}   & Frag\textcolor{teal}{$\downarrow$}\\
\midrule
ByteTrack       & 70.11 & 83.10 & 70.83 & 85.56 & 86.57  & 1380 & 974  & 44.59 & 47.57 & \underline{50.78}  & 57.64  & \underline{76.41} & \underline{37}     & \underline{193} \\
FastTracker       & 70.99 & 80.88 & 72.58 & 85.23 & 84.09  & 1342 & \underline{871}  & \textbf{46.58} & \textbf{48.33} & 50.13  & \underline{60.10}  & 68.98 & 56     & \textbf{130} \\
\hline
AgriSORT        & 57.19 & 46.99 & 60.63 & 70.35 & 59.90  & 1411 & \textbf{870}  & 43.90 & 30.13 & 46.09 & 55.00  & 50.68 & 82 & 236 \\
PineSORT        & 70.73 & 78.25 & 70.35 & 85.97 & 85.78 & 4640 & 1163 & 45.81 & 44.90 & 46.44 & \textbf{61.33} & 72.44 & 703 & 312 \\
NTrack          & \textbf{72.25} & \textbf{84.13} & \textbf{74.81} & \textbf{89.95} & \underline{90.74}  & \underline{1028} & 920  & 24.74 & 39.37 & 17.08  & 29.62 & 40.39 & 310 & 343 \\
\hline
CropTrack       & \underline{72.17} & \underline{84.03} & \underline{74.68} & \underline{89.86} & \textbf{90.88} & \textbf{1020} & 897  & \underline{46.04} & \underline{48.30} & \textbf{53.35} & 59.69  & \textbf{79.05} & \textbf{23} & 207 \\
\bottomrule
\end{tabularx}
\caption{A comparison of state-of-the-art MOT methods on the TexCot22
\cite{T8/5M9NCI_2024} and AgriSORT-Grapes \cite{saraceni2024agrisort} datasets.}
\label{table:tracker_benchmarks}
\end{table*}

\begin{figure*}
\centering
\resizebox{\textwidth}{!}{\input{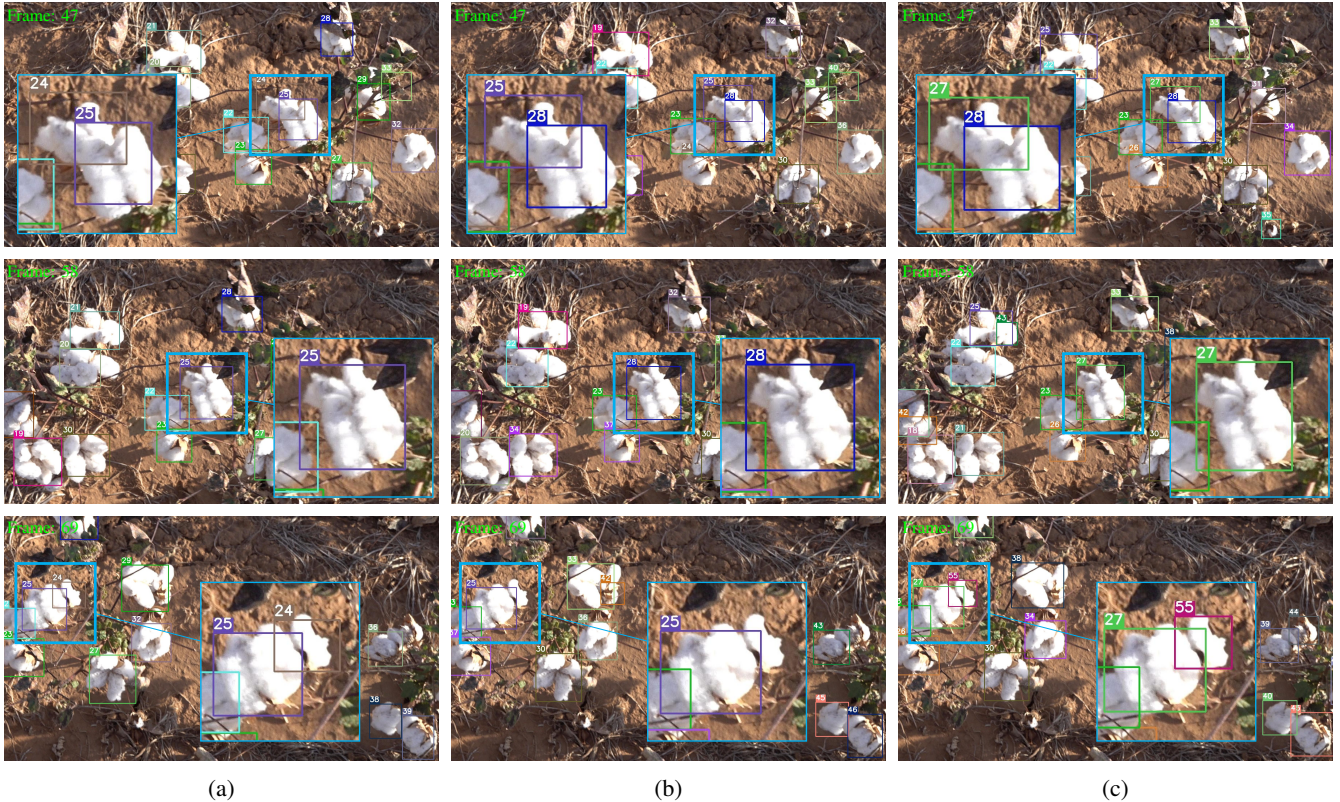}}
\caption{A qualitative comparison of (a) CropTrack, (b) NTrack, and (c) PineSORT
on a test sequence from TexCot22 \cite{T8/5M9NCI_2024} across multiple time
steps. The top row displays the tracking results at frame 47, while the middle
and bottom rows show the corresponding results at frame 58 and 69, respectively.
Each bounding box represents a tracklet and the color signifies its object ID. A
single color is used for the same object, while different colors are employed
for distinct objects. CropTrack yields superior ID preservation in cases of severe
occlusions, accurately restoring ID 24.}
\label{fig:qualitative_comparison}
\vspace{-3mm}
\end{figure*}

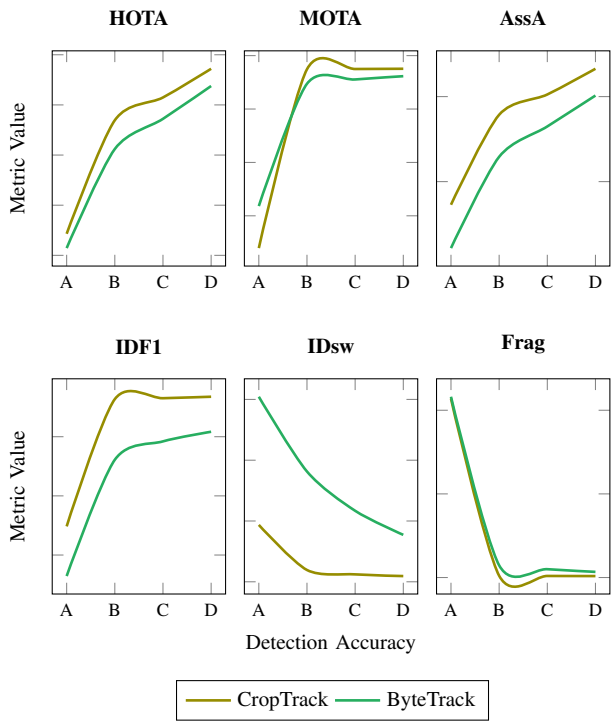
\begin{figure}
\centering
\begin{tikzpicture}
  \begin{groupplot}[
    group style={
      group size=3 by 2,   
      vertical sep=1.5cm,
      horizontal sep=.25cm,
      ylabels at=edge left,
      y descriptions at=edge left, 
      x descriptions at=edge bottom
    },
    label style={font=\footnotesize},
    title style={font=\bfseries\footnotesize},every axis plot/.append style={smooth, mark=.},
    width=0.45\linewidth,
    height=0.25\textwidth,
    xtick=data,
    xticklabel style={rotate=0, anchor=north, font=\scriptsize},
    tick label style={font=\scriptsize},
    ylabel={Metric Value},
    yticklabel=\empty,
    legend style={at={(-.6,-.4)}, anchor=north, legend columns=-1, font=\footnotesize}
  ]
  \nextgroupplot[title={HOTA}, symbolic x coords={A,B,C,D}]
  \addplot[line width=1pt, style={olive}]   coordinates {(A,54.296) (B,76.941) (C,81.484)  (D,87.209) };
  \addplot[line width=1pt, style={ggreen}] coordinates {(A,51.439) (B,71.207 ) (C,77.222) (D, 83.771)};

  \nextgroupplot[title={MOTA}, symbolic x coords={A,B,C,D}]
  \addplot[line width=1pt, style={olive}]   coordinates {(A, 63.95) (B,97.447) (C,97.496) (D,97.54)};
  \addplot[line width=1pt, style={ggreen}]  coordinates {(A, 71.82) (B,94.73) (C,95.538 ) (D,96.15)};
  
  \nextgroupplot[title={AssA}, symbolic x coords={A,B,C,D}]
  \addplot[line width=1pt, style={olive}]  coordinates {(A, 54.558) (B,75.728) (C,80.501) (D, 86.552)};
  \addplot[line width=1pt, style={ggreen}] coordinates {(A,44.358) (B,65.836) (C,72.979) (D,80.252 )};

  \nextgroupplot[title={IDF1}, symbolic x coords={A,B,C,D}]
  \addplot[line width=1pt, style={olive}]  coordinates {(A,74.843) (B,96.355) (C,96.511)(D,96.761)};
  \addplot[line width=1pt, style={ggreen}] coordinates {(A,66.442) (B,86.107) (C,89.213) (D,90.839)};

  \nextgroupplot[title={IDsw}, symbolic x coords={A,B,C,D}, xlabel={Detection Accuracy}]
  \addplot[line width=1pt, style={olive}]  coordinates {(A,1468) (B,1097) (C,1061) (D,1047)};
  \addplot[line width=1pt, style={ggreen}] coordinates {(A,2522) (B,1907) (C,1584) (D,1386)};

  \nextgroupplot[title={Frag}, symbolic x coords={A,B,C,D}]
  \addplot[line width=1pt, style={olive}]  coordinates {(A,10700) (B,109) (C,95) (D,88)};
  \addplot[line width=1pt, style={ggreen}] coordinates {(A,10808) (B,731) (C,499) (D, 339)};
  \legend{CropTrack, ByteTrack}
  \end{groupplot}
\end{tikzpicture}
\caption{Tracking performance under varying detection accuracies.}
\label{fig:varying_detection_accuracy}
\end{figure}

\begin{table}
\centering
\begin{tabular}{c|ccc}
\hline
\textbf{Noise Level} & LN probability & FN rate & FP rate \\
\hline
A & 0.4 & 0.2 & 0.2 \\
B & 0.4 & 0.0 & 0.0 \\
C & 0.2 & 0.0 & 0.0 \\
D & 0.0 & 0.0 & 0.0 \\
\hline
\end{tabular}
\caption{The noise level settings for the detection perturbation experiments.}
\label{tab:noise_level_settings}
\end{table}

\begin{table*}
\centering
\begin{tabular}{lccccccc}
\toprule
  & HOTA\textcolor{teal}{$\uparrow$} & MOTA\textcolor{teal}{$\uparrow$} & AssA\textcolor{teal}{$\uparrow$} & IDF1\textcolor{teal}{$\uparrow$} & IDP\textcolor{teal}{$\uparrow$} & IDsw\textcolor{teal}{$\downarrow$} & Frag\textcolor{teal}{$\downarrow$} \\
  \midrule
  ByteTrack  & 70.11             & 83.10             & 70.83             & 85.56                          & 86.57             & 1380             & \underline{974} \\
  +Re-ID     & 52.05             & 74.18             & 42.15             & 56.71                          & 57.06             & 1349             & 1301 \\
  +Reranking & 71.42             & 83.57             & 73.35             & 88.49                          & 89.20             & 1129             & 989 \\
  +Greedy    & \underline{71.94} & \underline{83.79} & \underline{74.36} & \underline{89.41}  & \underline{90.09} & \underline{1044} & 980 \\
  +NSA       & \textbf{72.17}    & \textbf{84.03}    & \textbf{74.68}    & \textbf{89.86}        & \textbf{90.88}    & \textbf{1020}    & \textbf{897} \\
  \bottomrule
\end{tabular}
\caption{An ablation study of the proposed association modules evaluated on the
TexCot22 \cite{T8/5M9NCI_2024} test sequences.}
\label{tab:ablation_study}
\end{table*}

\begin{table*}
\centering
\begin{tabular}{lccccccc}
\toprule
  & HOTA\textcolor{teal}{$\uparrow$} & MOTA\textcolor{teal}{$\uparrow$} & AssA\textcolor{teal}{$\uparrow$} & IDF1\textcolor{teal}{$\uparrow$} & IDP\textcolor{teal}{$\uparrow$} & IDsw\textcolor{teal}{$\downarrow$} & Frag\textcolor{teal}{$\downarrow$} \\
  \midrule
CropTrack (Hungarian) & {44.57}        & \textbf{48.38} & {50.12}        & {57.72}        & {76.42}        & {41}        & {190} \\
\midrule
CropTrack (Ours)      & \textbf{46.04} & {48.30}        & \textbf{53.35} & \textbf{59.69} & \textbf{79.05} & \textbf{23} & \textbf{190} \\
\bottomrule
\end{tabular}
\caption{A comparison of the Hungarian matching and our proposed association on
the AgriSORT-Grapes \cite{saraceni2024agrisort} dataset.}
\label{tab:association_ablation}
\end{table*}

\subsection{Datasets}
\label{subsec:datasets}
We conducted experiments on the following agricultural MOT datasets: TexCot22
\cite{T8/5M9NCI_2024} and the table grapes dataset presented in AgriSORT
\cite{saraceni2024agrisort}, which we refer to as AgriSORT-Grapes. TexCot22
contains a total of 30 video sequences of which 13 are used for testing. The
sequences are 10 to 20 seconds long and capture cotton crop rows from the
overhead perspective at 4K resolution and varying frame rates. The dataset was
recorded at separate times of the day, accounting for varying illumination
conditions. AgriSORT-Grapes contains 4 video sequences, all of which are used
for testing. The dataset is composed of 10-second sequences at a resolution of
720p. The sequences are divided equally by frame rate, with half of the videos
recorded at 30 FPS and the remainder at 10 FPS. All sequences are recorded from
the side view while moving along the vineyard rows.

\subsection{Evaluation Metrics}
\label{subsec:evaluaton_metrics}
We used the TrackEval framework \cite{luiten2021hota}, which provides a
comprehensive set of metrics for MOT evaluation. Specifically, we reported
performance using higher-order tracking accuracy (HOTA), MOT accuracy (MOTA),
association accuracy (AssA), identification F1 score (IDF1), identification
precision (IDP), identity switches (IDsw), and fragmentations (Frag). HOTA
provides a balanced assessment by jointly capturing detection, association, and
localization quality, thereby offering a holistic measure of MOT performance.
MOTA primarily reflects detection performance as it aggregates false positives,
false negatives, and identity switches. AssA quantifies a tracker's ability to
maintain identity consistency across frames, which is vital for applications
such as crop yield estimation. IDF1 emphasizes identity preservation by
measuring the consistency of correctly identified detections over time. IDP
measures the accuracy of positive identifications made by system. IDsw
quantifies errors in identity assignment, whereas Frag measures interruptions in
continuous trajectories. Collectively, these metrics provide a detailed
evaluation of CropTrack's ability to track and re-identify crop instances over
time.

\subsection{Implementation Details}
\label{subsec:implementation_details}
CropTrack's appearance features, $(f \in \mathbb{R}^{1024})$, are extracted from
detections via the PHA \cite{zhang2023pha} model, which is pretrained on the
Market-1501 \cite{zheng2015scalable} dataset for person Re-ID. Unless otherwise
specified, the detection score threshold $\tau$ is set to 0.6. In
\eqref{eq:refined_pairwise_distance}, the maximum allowable neighbor distance
$\delta$ for potential association is set at 600, accommodating larger track
drifts. During the assignment step, associations are discarded if the IoU
between a detection box and a tracklet box falls below 0.2. Lost tracklets are
retained for up to 30 frames to account for potential reappearance. To ensure a
fair evaluation, we adopted the detection bounding boxes provided with the
dataset across all baseline methods. 

\subsection{Tracker Comparison}
\label{subsec:tracker_comparision}
As demonstrated in Table~\ref{table:tracker_benchmarks}, CropTrack establishes
superior identity preservation when evaluated against state-of-the-art trackers
on the TexCot22 and AgriSORT-Grapes datasets. It consistently outperforms all
other methods in identity-specific metrics, achieving the highest IDP and the
lowest IDsw on both benchmarks. Furthermore, CropTrack achieves competitive
results in all the main metrics (HOTA, MOTA, AssA, IDF1). Although CropTrack
excels in identity preservation, it does exhibit a higher number of fragmented
tracks. This is a direct consequence of its ability to accurately re-associate
and reactivate tracks after an occlusion, as illustrated in
Fig.~\ref{fig:qualitative_comparison}. The high tracking accuracy and reliable
identity preservation across two distinct crop types validates the effectiveness
of appearance-based association.

\subsection{Tracking Performance under Varying Detection Accuracy}
\label{subsec:tracking_performance_under_varying_detection_accuracy}
To confirm that our method performs well with current detectors and remains
robust as detection accuracy improves, we tested CropTrack under different
detection accuracy levels. Concretely, we introduced three types of
perturbations into the ground-truth bounding boxes.
\begin{enumerate}
  \item Localization noise (LN): We perturbed each bounding box to produce
  varying distortion levels by adding random spatial noise to its center
  coordinates and scaling noise to its width and height. 
  \item False negatives (FNs): We randomly removed a fraction of bounding boxes
  to simulate missed detections.
  \item False positives (FPs): We randomly sampled additional bounding boxes
  to represent spurious detections.
\end{enumerate}

The perturbation procedure was controlled by three parameters: LN probability,
FN rate, and FP rate. As shown in Fig.~\ref{fig:varying_detection_accuracy},
three distinct noise levels, labeled A, B, and C, were evaluated alongside a
ground-truth baseline, labeled D. The specific parameter configurations for each
noise level are detailed in Table~\ref{tab:noise_level_settings}. The results
demonstrate that CropTrack achieves considerably higher performance than
baseline approaches as detection quality increases. Notably, the gains in HOTA,
MOTA, AssA, and IDF1 indicate that our approach is more effective at maintaining
accurate and consistent object identities. Additionally, lower values of IDsw
and Frag reflect the robustness of our method in minimizing identity switches
and fragmentation. Collectively, these findings highlight the resilience of
CropTrack to variations in detection quality and its superior capacity to
maintain long-term object trajectories.

\subsection{Ablation Study}
\label{subsec:ablation_study}
The development of CropTrack is detailed via an ablation study shown in
Table~\ref{tab:ablation_study}. ByteTrack is employed as the baseline and
appearance-based Re-ID is introduced for unmatched detections after the first
motion-based association. Choosing cosine similarity between detection features
and EMA feature prototypes results in a drastic reduction in performance. This
is expected as the model is originally trained for person Re-ID and applied in a
zero-shot manner with no fine-tuning. However, swapping cosine similarity with
our reranking distance results in a major improvement across all metrics,
excluding fragmentation. Leveraging the reranking distance, we replace the first
motion-based association with our one-to-many association with greedy
appearance-based conflict resolution. Lastly, we form CropTrack by replacing the
vanilla Kalman filter with the NSA Kalman filter.

To further validate our association strategy, we conducted an additional
ablation study on the AgriSORT-Grapes dataset. Specifically, we compare the full
CropTrack model against a baseline utilizing only Hungarian-based matching.
Compared to the baseline, the results in Table~\ref{tab:association_ablation}
show that CropTrack significantly improves HOTA, AssA, IDF1, IDP, and IDsw. This
indicates that appearance-based features are beneficial for identity
preservation in agricultural tracking.

\section{Conclusion}
\label{sec:conclusion}
This letter presented CropTrack, a MOT framework for infield crop monitoring
based on Re-ID. CropTrack integrates reranking-enhanced appearance association,
a one-to-many association with appearance-based conflict resolution strategy,
and an EMA-based prototype feature bank. These components collectively enhance
robustness under occlusion, dense plant structures, and high visual similarity.
Evaluations on the TexCot22 and AgriSORT-Grapes datasets, which serve as
benchmarks for agricultural tracking, show that CropTrack achieves
state-of-the-art performance in tracking accuracy and identity consistency.
Future research will investigate multimodal sensor fusion and online adaptation
to further generalize tracking across diverse environmental conditions.

\bibliographystyle{IEEEtran}
\bibliography{croptrack_a_tracking_with_re-identification_framework_for_precision_agriculture}

\end{document}